\documentclass{article}

\newcommand{\gp}{G_{\text{pos}}}
\newcommand{\gn}{G_{\text{neg}}}
\newcommand{\lps}{L_{\text{pos}}}
\newcommand{\lng}{L_{\text{neg}}}
\newcommand{\lff}{L_{\text{FF}}}
\newcommand{\lfe}{L_{\text{SymBa}}}
\newcommand{\cc}{\cellcolor[HTML]{EFEFEF}}

\usepackage[square,numbers]{natbib}
\usepackage[preprint]{neurips_2022}

\usepackage[ruled,vlined]{algorithm2e}
\usepackage{times}
\usepackage{multirow}
\usepackage{epsfig}
\usepackage{graphicx}
\usepackage{amsmath}
\usepackage{amssymb}
\usepackage{chngpage}
\usepackage{mathtools}
\usepackage{caption}
\usepackage[table,xcdraw]{xcolor}
\usepackage{multirow}
\usepackage{listings}
\usepackage{subcaption}
\usepackage{sidecap}

\usepackage[utf8]{inputenc} 
\usepackage[T1]{fontenc}    
\usepackage{hyperref}       
\usepackage{url}            
\usepackage{booktabs}       
\usepackage{amsfonts}       
\usepackage{nicefrac}       
\usepackage{microtype}      
\usepackage{xcolor}         
\usepackage{amsmath}
\usepackage{sidecap}
\sidecaptionvpos{figure}{c}

\definecolor{textblue}{rgb}{.2,.2,.7}
\definecolor{textred}{rgb}{0.54,0,0}
\definecolor{textgreen}{rgb}{0,0.43,0}
\usepackage{listings}
\title{SymBa: Symmetric Backpropagation-Free Contrastive Learning with Forward-Forward Algorithm for Optimizing Convergence}

\author{
  Heung-Chang Lee\thanks{Equal contribution}\, $^\dagger$, Jeonggeun Song$^\ast$\thanks{Corresponding authors}\\
  Kakao Enterprise\\
  Seongnam-si, Republic of Korea\\
  \texttt{andrew.com@kakaoenterprise.com; po.ai@kakaoenterprise.com} \\
}

\begin{document}

\maketitle

\begin{abstract}
  The paper proposes a new algorithm called SymBa that aims to achieve more biologically plausible learning than Back-Propagation (BP). The algorithm is based on the Forward-Forward (FF) algorithm, which is a BP-free method for training neural networks. SymBa improves the FF algorithm's convergence behavior by addressing the problem of asymmetric gradients caused by conflicting converging directions for positive and negative samples. The algorithm balances positive and negative losses to enhance performance and convergence speed. Furthermore, it modifies the FF algorithm by adding Intrinsic Class Pattern (ICP) containing class information to prevent the loss of class information during training. The proposed algorithm has the potential to improve our understanding of how the brain learns and processes information and to develop more effective and efficient artificial intelligence systems. The paper presents experimental results that demonstrate the effectiveness of SymBa algorithm compared to the FF algorithm and BP.
\end{abstract}

\section{Introduction}
\label{introduction}
In recent years, deep learning has made remarkable strides in various research domains. By utilizing stochastic gradient descent with a large number of parameters and abundant data, deep learning has achieved state-of-the-art results. This success has sparked interest in investigating the learning mechanisms employed by biological systems, particularly the brain. Researchers are curious to explore whether the mechanisms used in deep learning have any resemblance to those employed in the brain. However, despite significant efforts, the Back-Propagation (BP) algorithm~\cite{rumelhart1986learning}, which is a commonly used technique in deep learning, is still not considered a plausible model for how the cortex learns.

This has motivated researchers to explore alternative theories of how the brain learns and processes information. Recent research has focused on developing biologically plausible neural networks that can learn in a manner similar to the brain. These models have the potential to improve our understanding of how the brain works and to develop more effective and efficient artificial intelligence systems.

Our paper introduces SymBa algorithm, which achieves more human-like training by avoiding BP. Based on the preprint of the Forward-Forward (FF) algorithm ~\cite{hinton2022forward}, SymBa algorithm stabilizes converging behaviors and improves overall performance.
\newpage
The contributions of our algorithm can be summarized as follows.

\begin{itemize}
    \item The equations of FF algorithm are difficult to converge towards the global minima due to conflicting converging directions for loss of positive and negative samples. Our approach addresses this problem by ensuring that both gradients converge in the same direction, resulting in improved and efficient convergence during training.
    \item To accomplish the classification task and apply the overlay, the previous FF Algorithm conducted one-hot encoding of class information on the input picture. However, this approach has a disadvantage in that it can cause the class information to be lost during training, resulting in poor performance. To address this issue, we modify the FF algorithm by introducing Intrinsic Class Pattern (ICP) containing class information behind each channel. This modification prevents the class information from being lost during training, improving the overall performance of the algorithm.
\end{itemize}

\begin{figure}[t]
    \centering
    \includegraphics[width=\linewidth]{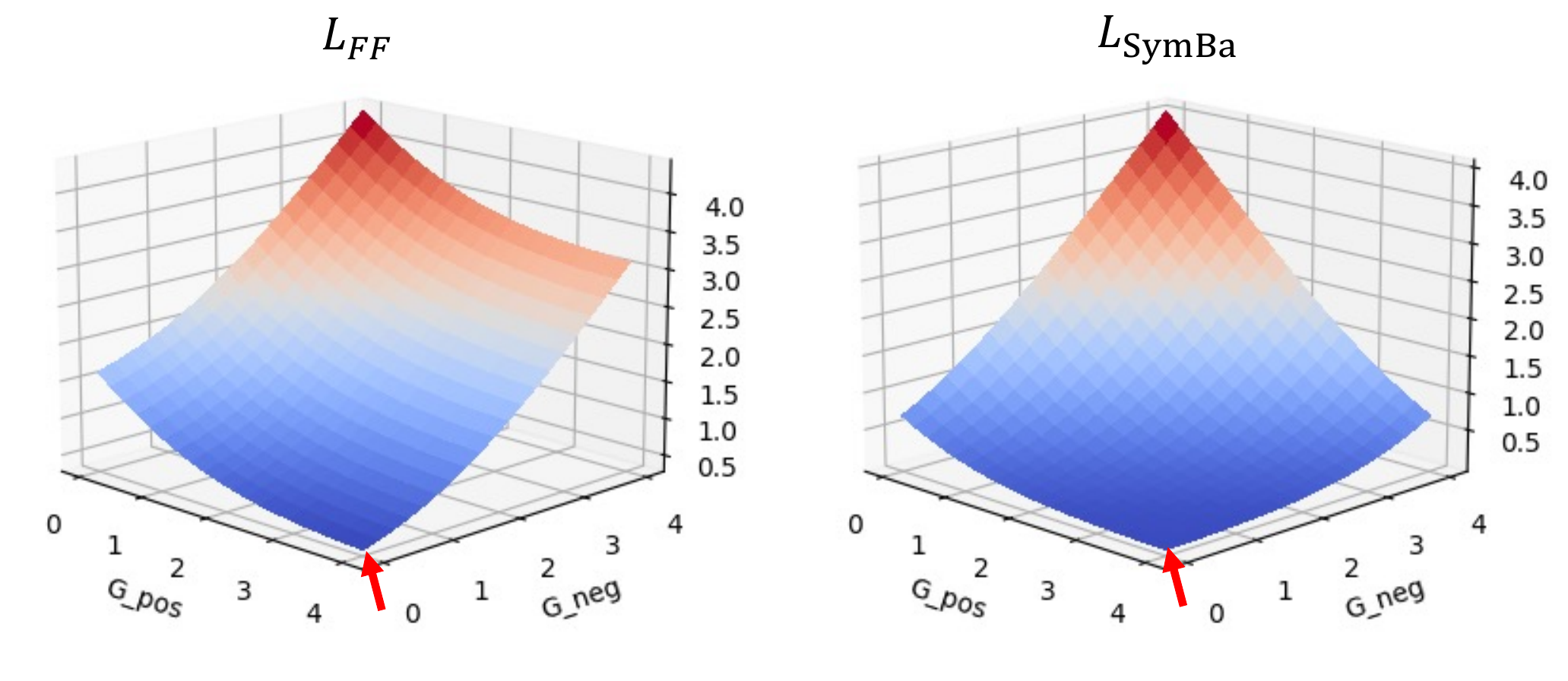}
    \caption{\textbf{Comparison of} $\lff$ \textbf{and} $\lfe$ \textbf{over} $\gp$ \textbf{and} $\gn$\textbf{.} The direction of convergence is that $\gp$ increases continuously, while $\gn$ decreases to 0. In order to clarify, the red arrows in the figure represent the points of convergence. In the case of $\lff$, there is a significant gap of gradient scales depending on the choice of initial points. Conversely, $\lfe$ exhibits precise symmetry across all the values of $\gp$ and $\gn$. In contrast to $\lff$, $\lfe$ converges along the same slopes in most cases regardless of the choice of initial point.}
    \label{fig:loss_comparison}
\end{figure}

\section{Related Works}
\label{related_works}

There has been a significant amount of research on developing biologically plausible neural networks that can learn in a manner similar to the brain. One such approach is the Hebbian learning rule, which was from Donald Hebb's 1949 book \cite{hebb1949organization}. Hebb proposed that when two neurons are repeatedly activated at the same time, the strength of the connection between them increases. This idea, known as Hebb's rule, has since been widely studied and extended in many different contexts.
Another approach is spike-timing-dependent plasticity (STDP), which modifies the synaptic strengths between neurons based on the relative timing of their spikes \cite{gerstner2002spiking}. STDP has been used to train spiking neural networks, which have been shown to be computationally efficient and capable of solving a wide range of tasks \cite{tavanaei2019deep}.

Additionally, the popular method for estimating the parameters of a probabilistic model is Noise Contrastive Estimation (NCE), which was introduced by Gutmann and Hyvärinen \cite{gutmann2010noise}. NCE has several advantages, including its simplicity and efficiency, as it only requires the computation of simple logistic regression. It has been applied in a variety of contexts, including learning binary codes for image retrieval \cite{mnih2012learning}, training deep neural networks for high-dimensional data modeling \cite{choromanska2015logarithmic}. However, NCE also has some limitations, such as the requirement for an explicit noise distribution and sensitivity to the quality of the noise distribution. 

Recent work has also focused on developing biologically plausible activation functions for neural networks. One such function is the rectified linear unit (ReLU), which has been shown to be more biologically plausible than traditional sigmoid activation functions \cite{glorot2011deep}. Despite the success of these biologically inspired approaches, the Back-Propagation (BP) algorithm, which is commonly used in deep learning, is still not considered a plausible model for how the cortex learns. This has motivated researchers to explore alternative theories of how the brain learns and processes information. In this context, the Forward-Forward (FF) algorithm has been proposed as a more biologically plausible alternative to BP. The FF algorithm avoids using BP by training each layer independently to maximize the goodness of positive samples and minimize that of negative samples. Our paper builds on the FF algorithm and introduces SymBa algorithm, which improves the overall performance of the FF algorithm by balancing the positive and negative losses during training. This loss of FF algorithm is similar to the contrastive loss. \cite{chen2020simple, he2020momentum, grill2020bootstrap}

\begin{figure}
    \centering
    \includegraphics[width=0.9\linewidth]{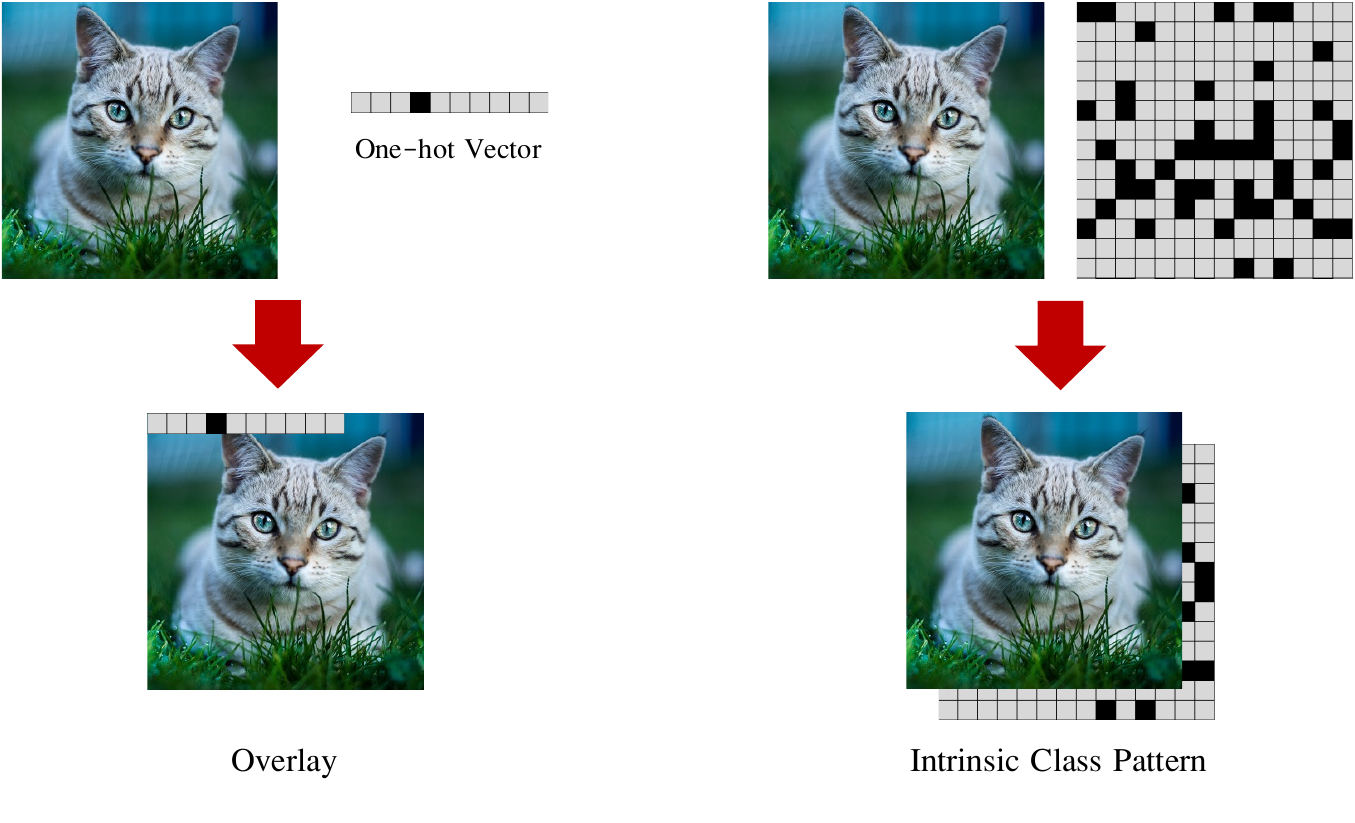}
    \caption{\textbf{Intrinsic class pattern.} When labels are overlaid directly onto input images, there is a significant loss of representations. (i.e. The edge of the cat's right ear in the above image.) In contrast, intrinsic class patterns, which are randomly generated unique fixed patterns for each class, allow the model to recognize the labels by identifying the corresponding class pattern. By avoiding the loss of input pixels, intrinsic class patterns provide more useful information about the classes without wasting the model's capacity to conjecture obscured parts of the input.}
    \label{fig:intrinsic_class_pattern}
\end{figure}

\section{Method}
\label{method}
The FF algorithm does not rely on BP to emulate the function of the human brain; instead, each layer is trained independently to maximize the goodness of positive samples and minimize that of negative samples, while the computational graph does not connect any of the layers. 

Nevertheless, in the original implementation, these samples frequently cause asymmetric gradients, which result in suboptimal performance and delayed convergence. To improve the algorithm's performance, it is necessary to balance the positive and negative losses during training. It has been demonstrated that balancing the losses greatly enhances the efficiency and convergence speed of the FF algorithm. 

\begin{figure}[htb!]
     \centering
     \includegraphics[width=0.48\linewidth]{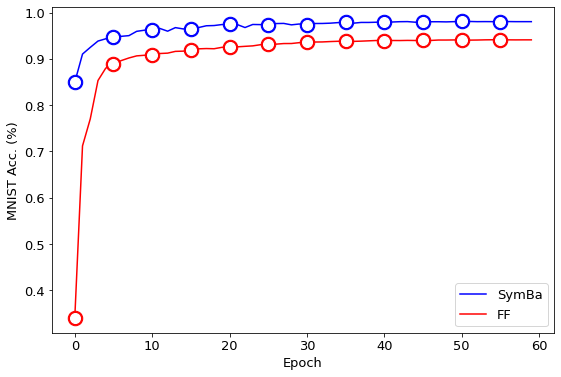}
     \includegraphics[width=0.48\linewidth]{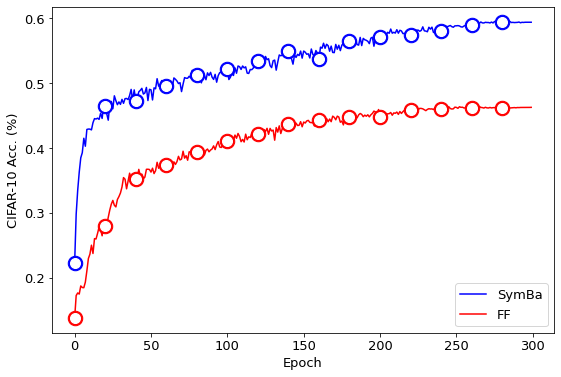}
     \caption{\textbf{The measurement of converging speed.} The graph depicts the change in the accuracy of FF and SymBa by epochs. The accuracy converges quickly in both cases, but SymBa does so considerably more rapidly than the FF. In particular, SymBa method significantly outperforms the FF algorithm in the MNIST dataset~\cite{lecun1998gradient} 0 epoch, and it continuously surpasses the FF algorithm in the CIFAR-10 dataset~\cite{krizhevsky2009learning}.}
     \label{fig:converging_speed}
\end{figure}


\subsection{Imbalance of Positive-Negative Losses}
The previous implementation of the FF algorithm utilizes the Noise Contrastive Estimate (NCE) loss function. The original loss function of the FF algorithm is as follows.
\begin{align}\label{eq:NCL}
    &\gp=\sum_{j}y^2_{\text{pos},j}, \gn=\sum_{j}y^2_{\text{neg},j}\\
    &\lps=\log{(1+e^{\theta-\gp})},
    \lng=\log{(1+e^{\gn-\theta})}\\
    &L_{\text{FF}}=\lps+\lng
\end{align}
Note that $\theta$ is threshold. When the domain of the goodness $G$ is the set of all real numbers, the above formula is symmetric about $G$. However, since $G$ must be greater than 0, the positive and negative losses have different convergence behaviors, leading to imbalanced gradients during the training and slowing down the convergence.
\begin{align}\label{eq:pos_neg_grad}
    &\nabla \lps=-\cfrac{2e^{\theta-G_\text{pos}}}{1+e^{\theta-\gp}}\sum_{j}y_{\text{pos},j}\nabla y_{\text{pos},j}\\
    &\nabla \lng=\cfrac{2e^{G_\text{neg}-\theta}}{1+e^{\gn-\theta}}\sum_{j}y_{\text{neg},j}\nabla y_{\text{neg},j}
\end{align}
The gradients for positive and negative losses can be expressed as Eq 
\eqref{eq:pos_neg_grad}. Due to the condition that $G>0$, $\gp \to \infty$ and $\gn \to 0$ as the training progresses. Consider $\epsilon$ and $\Omega$ as sufficiently small and large numbers, respectively. Substituting them to $\gn$ and $\gp$ each, we can observe varying outcomes in the vicinity of the convergence point.
\begin{align}\label{eq:pos_neg_grad_range}
    \nabla \lps&=-\cfrac{2e^{\theta-\Omega}}{1+e^{\theta-\Omega}}\sum_{j}y_{\text{pos},j}\nabla y_{\text{pos},j}\approx 0\\
    \nabla \lng&=\cfrac{2e^{\epsilon-\theta}}{1+e^{\epsilon-\theta}}\sum_{j}y_{\text{neg},j}\nabla y_{\text{neg},j}\approx \cfrac{2e^{-\theta}}{1+e^{-\theta}}\sum_{j}y_{\text{neg},j}\nabla y_{\text{neg},j}
\end{align}
It shows that the gradient for positive and negative samples behave differently, toward the end of the training process. Furthermore, the final convergence values of positive and negative losses are different. The presence of this discrepancy impedes the model's ability to attain global minima.
\begin{align}\label{eq:convergence_values}
    \gp \to \infty& \Leftrightarrow \lps \to 0 \\
    \gn \to 0& \Leftrightarrow \lng \to \log{(1+e^{-\theta})}
\end{align}
In order to obtain the proper convergence properties, we introduce an alternative algorithm, SymBa.

\subsection{Balanced Contrastive Loss for Equilibrium of Positive-Negative Losses}
To address the asymmetric nature of the original implementation, we propose an alternative loss function as follows.
\begin{align}\label{eq:delta_loss}
    &\Delta=\gp-\gn\\
    &\lfe=\log{(1+e^{-\alpha\Delta})}
\end{align}
where $\alpha$ is a simple scale factor. Our new loss function has various benefits for enhancing the stability of training. To begin with, it eliminates the requirement to consider the equilibrium between $\gp$ and $\gn$. As $\lfe$ solely depends on the discrepancy between two losses, it is inherently symmetric.
\begin{align}\label{eq:ff_eq_convergence_values}
    \gp \to \infty, \gn \to 0 \Leftrightarrow \lfe \to 0
\end{align}
Furthermore, it utilizes the explicit relation of $\gp$ and $\gn$. It enables the model to discern the correlation between the two quantities along the batch dimension. It facilitates the model to infer the association between two sets of samples.

For quantitative analysis, we conducted experiments on the MNIST, CIFAR-10, and CIFAR-100 datasets. The results showed that our proposed algorithm outperformed the existing FF and BP algorithms on  all datasets we experimented on. Moreover, it exhibited significantly better performance in terms of convergence speed than the existing FF algorithm. While the previous algorithm converges towards the end of the training, our algorithm rapidly approached the optimal performance after only a few initial epochs.

\renewcommand{\arraystretch}{1.2} 
\begin{table}[t]
    \centering
    \caption{\textbf{The results on MNIST.} This table indicates the result of experiments using the BP, FF, and SymBa algorithms while varying the number of layers and channels on the MNIST dataset. The labeling method and hyper-parameters suitable for each algorithm were used. Forthermore, the best test error values are bolded, only the outcomes of our suggested SymBa algorithm are shaded} 
    \begin{tabular}{c|c|l|l|l|c}
    \hline
        \rowcolor[HTML]{EFEFEF}
        \#Layers & \#Channels & Algorithm & Labeling Method & Details & Test Error(\%) \\ \hline
        \multirow{6}{*}{2} & \multirow{3}{*}{500} & BP & None & & 1.74 \\ \cline{3-6}
        ~ & ~ & FF & Overlay & $\theta=2.0$ & 6.10 \\ \cline{3-6}
        ~ & ~ & $\cc$SymBa & $\cc$ICP & $\cc$$\alpha=4.0$ & $\cc$\textbf{1.65} \\ \cline{2-6}
        ~ & \multirow{3}{*}{2000} & BP & None & & 1.56 \\ \cline{3-6}
        ~ & ~ & FF & Overlay & $\theta=2.0$ & 6.93 \\ \cline{3-6}
        ~ & ~ & $\cc$SymBa & $\cc$ICP & $\cc$$\alpha=4.0$ & $\cc$\textbf{1.52} \\ \hline
        \multirow{6}{*}{3} & \multirow{3}{*}{500} & BP & None & & \textbf{1.77} \\ \cline{3-6}
        ~ & ~ & FF & Overlay & $\theta=2.0$ & 5.79 \\ \cline{3-6}
        ~ & ~ & $\cc$SymBa & $\cc$ICP & $\cc$$\alpha=4.0$ & $\cc$\textbf{1.77} \\ \cline{2-6}
        ~ & \multirow{3}{*}{2000} & BP & None & & 1.58 \\ \cline{3-6}
        ~ & ~ & FF & Overlay & $\theta=2.0$ & 6.59 \\ \cline{3-6}
        ~ & ~ & $\cc$SymBa & $\cc$ICP & $\cc$$\alpha=4.0$ & $\cc$\textbf{1.42} \\ \hline
    \end{tabular}
    \label{tb:mnist}
\end{table}

\subsection{Intrinsic Class Patterns as Labels}\label{icp}
Since the FF algorithm employs contrastive loss, a fine-tuning process is required using the BP algorithm to evaluate its performance. In the original paper, the label information is directly injected into the input images by overlaying one-hot encoding. Rather than relying on a classifier, accuracy can be evaluated by measuring the goodness of each label and selecting the label with the highest score, as determined by $\text{argmax}_y G(x, y)$. This approach is both clever and valid, as it avoids the need to utilize BP during training.

However, the input information is compromised by overlaying the labels. The one-hot encoding representing the labels partially obscure the input images. Since the size of the one-hot encoding is determined by the number of classes, a significant portion of the images is removed as the number of classes increases. For instance, in CIFAR-100 dataset with 100 classes, each label covers approximately 9.77\% of the image. It is much higher than the 1.27\% coverage of labels in MNIST dataset so cannot be neglected. Furthermore, even if the number of classes is negligibly small, it cannot be guaranteed that its negative impact is negligible as well.

To address this problem, we propose an alternative method to incorporating class labels into input images, which is called \textbf{Intrinsic Class Patterns}. Rather than overlaying one-hot encoding, we generate a non-trainable random discrete pattern for each class, which is concatenated to input channels. This approach avoids directly covering images, thus preserving the original input information.

In our experiments, incorporating intrinsic class patterns in the inputs results in improved performance for both the original FF algorithm and our proposed algorithm on the CIFAR-10 and CIFAR-100 datasets.

\begin{table}[t]
    \centering
    \caption{\textbf{The results on CIFAR-10 and 100.} These experiments are the results of the BP, FF, and SymBa algorithms on the CIFAR-10 and 100 datasets by adjusting the number of channels. Since CIFAR datasets contain more diverse images, three fully-connected layers are used for all cases to obtain enough model capacity. For a fair comparison, the whole experiments are conducted in the exact same environment as the MNIST experiment, and the best-performing results are bolded and the SymBa algorithm is colored to increase visibility.}
    \begin{tabular}{c|c|l|l|l|c}
    \hline
        \rowcolor[HTML]{EFEFEF}
        Dataset & \#Channels & Algorithm & Labeling Method & Details & Test Error(\%) \\ \hline
        \multirow{6}{*}{CIFAR-10} & \multirow{3}{*}{2000} & BP & None & & 42.66 \\ \cline{3-6}
        ~ & ~ & FF & Overlay & $\theta=2.0$ & 49.32 \\ \cline{3-6}
        ~ & ~ & $\cc$SymBa & $\cc$ICP & $\cc$$\alpha=4.0$ & $\cc$\textbf{41.23} \\ \cline{2-6}
        ~ & \multirow{3}{*}{3072} & BP & None & & 43.14 \\ \cline{3-6}
        ~ & ~ & FF & Overlay & $\theta=2.0$ & 49.63 \\ \cline{3-6}
        ~ & ~ & $\cc$SymBa & $\cc$ICP & $\cc$$\alpha=4.0$ & $\cc$\textbf{40.91} \\ \hline
        \multirow{3}{*}{CIFAR-100} & \multirow{3}{*}{3072} & BP & None & & 71.12 \\ \cline{3-6}
        ~ & ~ & FF & Overlay & $\theta=2.0$ & 81.85 \\ \cline{3-6}
        ~ & ~ & $\cc$SymBa & $\cc$ICP & $\cc$$\alpha=4.0$ & $\cc$\textbf{70.72} \\ \hline
    \end{tabular}
    \label{tb:cifar}
\end{table}

\section{Experiments}
\label{experiments}

As discussed in the method session, we want to conduct experiments in order to determine whether our SymBa method can outperform BP and converge more efficiently than FF on a variety of datasets, including MNIST, CIFAR-10, and 100.
Our experimental environment was based on the Classification Task for Supervised Learning.
The accuracy criterion for BP was if the highest value of the final Softmax previously employed corresponded to the correct answer class, while the accuracy criterion for FF and SymBa was whether the class with the most goodness used in FF corresponds to the correct answer class.

Several hyper-parameters were left unmentioned in the paper of FF, but we made our best effort to replicate FF.
Most of the hyper-parameters described in the original paper were utilized, while the unknown hyper-parameters were discovered via various experiments. Additionally, we conducted the whole experiment with essential settings which does not utilize regularization like weight decay\cite{loshchilov2017decoupled} or the drop-out method~\cite{srivastava2014dropout} to compare fairly. Nonetheless, we employed basic augmentation methods such as random cropping and random horizontal flipping in all experiments on CIFAR-10 and 100.

\subsection{Experiments with MNIST}
As described in the paper, we trained architectures with 2 or 3 fully connected layers on the MNIST dataset. Since the other hyperparameters were not given, we selected them manually by conducting several experiments. All models were trained using Adam optimizer\cite{kingma2014adam} for 120 epochs with the batch size of 4096, and we swept the learning rates in $\{0.001, 0.01\}$ range. The scale factor for $\lfe$, $\alpha$, did not affect the performance significantly when it was set to any number within the range of $\{1.0, 4.0\}$. It made the difference less than $0.1\%$ only, for the experiments on both MNIST and CIFAR datasets. Therefore, we reported the results of $\alpha=4.0$ cases for all experiments.

In our experiments, we observed that SymBa exhibits noticeably faster convergence than FF from the first epoch onwards. (See Figure~\ref{fig:converging_speed}.) Furthermore, The performance gap between SymBa and FF is maintained until the final epoch, resulting in higher final performance. We anticipate that it is closely related to the stabilization of the converging curve discussed above. As shown in Table~\ref{tb:mnist}, we experimented with BP, FF, and SymBa under various setups, and SymBa consistently achieved the best performance across all setups.

\subsection{Experiments with CIFAR-10 and 100}

In contrast to the MNIST dataset, CIFAR datasets contain more complex features and color information. Furthermore, for the case of CIFAR-100 dataset, it has a large number of classes and fewer data per class. Since the images in CIFAR datasets contain more detailed representations, we employed models with three fully-connected layers to ensure enough model capacity. For the other experimental setups, we utilized substantially identical configurations to those used in the MNIST experiments. As shown in Table~\ref{tb:cifar}, our experiments show that SymBa algorithm outperforms in all experiments, while FF algorithm shows the lowest performance. It claims that SymBa can be trained better than BP on more complicated tasks.

\subsection{Ablation Study}
\textbf{Ablation study between Overlay and ICP.} 
We conducted an ablation study on the FF and SymBa algorithm for overlaying one-hot encoding, which was employed in the current FF, and ICP to investigate the impact of ICP. The dataset was conducted on CIFAR-10, and the channel used in the experiment session was 3072. ICP outperformed the overlay for all algorithms, as indicated in Table ~\ref{tb:abl_icp}, while the SymBa algorithm outscored all FF algorithms even for the overlay.

\begin{table}[t]
    \centering
    \caption{\textbf{Ablation study between Overlay and ICP.} As mentioned in Sec.~\ref{icp}, both FF and SymBa achieved higher performance by replacing one-hot encodings with intrinsic class patterns. This substitution allowed for more effective learning, as the models could better capture the inherent characteristics of each class.}
    \begin{tabular}{c|c|l|l|c}
    \hline
        \rowcolor[HTML]{EFEFEF}
        Dataset & \#Channels & Algorithm & Labeling Method & Test Error(\%) \\ \hline
        \multirow{4}{*}{CIFAR-10} & \multirow{4}{*}{3072} & \multirow{2}{*}{FF} & Overlay & 49.63 \\ \cline{4-5}
        ~ & ~ & ~ & $\cc$ICP & $\cc$\textbf{48.83} \\ \cline{3-5}
        ~ & ~ & \multirow{2}{*}{SymBa} & Overlay & 41.23 \\ \cline{4-5}
        ~ & ~ & ~ & $\cc$ICP & $\cc$\textbf{40.91} \\ \hline
    \end{tabular}
    \label{tb:abl_icp}
\end{table}

\begin{figure}
    \centering
    \includegraphics[width=0.7\linewidth]{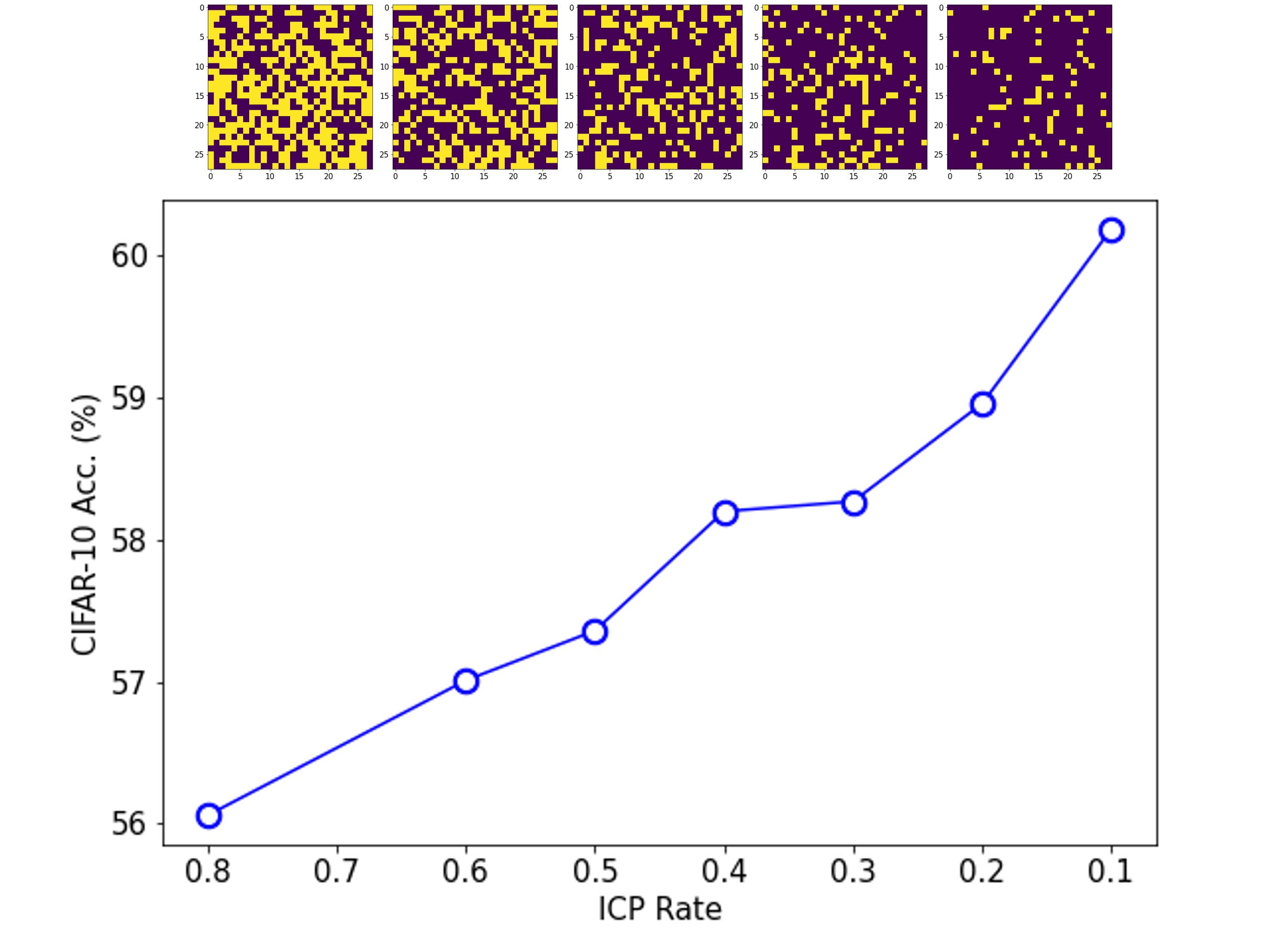}
    \caption{\textbf{Ablation study on ICP rate.} The class patterns are generated by selecting random pixels from black images and assigning them a value of 1. ICP rate refers to the sampling rate used to fill in these patterns, and our observations indicate that the class patterns with low ICP rates facilitate the models learning of class differences. As sparse patterns render the distinctions between classes more discernible, the final performance demonstrates a near-linear relationship with sparsity, as depicted in the graph.}
    \label{fig:abl_icp_rate}
\end{figure}

\textbf{Ablation study on ICP rate.}
The ablation study is conducted to find the optimal rate of ICP, which is one of the contributions of the paper. The values varied from 0.1 to 0.8, with 0.8 being a noisy input and 0.1 denoting a sparse one. The yellow dots in the upper pictures from Figure~\ref{fig:abl_icp_rate} are noise, and the purple dots are input. The graph in Figure~\ref{fig:abl_icp_rate} demonstrates that the accuracy of the CIFAR-10 improves as the noise level lowers. Nevertheless, if it is smaller than 0.1, the loss converges to NaN since it is impossible to extract class-specific information on its own. As a consequence, we established that 0.1 was the ideal rate for ICP and utilized this rate in all of our experiments.

\section{Discussion and conclusion}
\label{conclusion}
Back-propagation (BP) algorithm has been the de facto standard for training deep learning models for a long time. Despite its remarkable performance, several studies have suggested that it is not suitable for emulating the way the brain learns in reality. Additionally, BP requires significant computing power and vast amounts of data to learn knowledge of unknown domains, making it inefficient compared to the human brain, which can adapt flexibly to various areas with only a few samples.

Our proposed algorithm, SymBa, aims to mimic the learning process of the human cortex without relying on BP. It builds upon Forward-Forward (FF) algorithm, in which gradients are computed internally within each layer by estimating the contrastive loss between two forward passes of positive and negative inputs, rather than propagating information between layers. Although FF algorithm was a promising approach, issues arose, such as imbalanced losses near convergence and the loss of input information due to the overlaying of one-hot encodings on input images for evaluation without fine-tuning.

In SymBa, we address these limitations by combining the positive and negative losses into a unified loss, achieving a perfect balance while preserving the contrastive properties of the original algorithm. Additionally, we concatenate intrinsic noise patterns for each class rather than obscuring input images, thus conserving their representations. As a result, SymBa outperforms the BP algorithm on widely-used benchmark datasets, such as MNIST, CIFAR-10, and CIFAR-100.

As an extension of various approaches to reproduce the behaviors of the brain, the results of SymBa outperform those of BP with a significant margin. However, BP has been widely used in a vast number of tasks, while there exist many different tasks for SymBa to prove its ability. While SymBa is not be a perfect replacement for BP, there is potential for further development towards pioneering more human-like algorithms than BP.

\bibliography{neurips}

\end{document}